\documentclass{bmvc2k}

\usepackage{amsfonts}
\usepackage{amsmath}
\usepackage{graphicx}
\usepackage{pifont}% http://ctan.org/pkg/pifont
\usepackage{hyperref}

\newcommand{\cmark}{\ding{51}}%
\newcommand{\xmark}{\ding{55}}%
\newcommand{\bx}{\mathbf{x}}
\newcommand{\by}{\mathbf{y}}
\newcommand{\bz}{\mathbf{z}}
\newcommand{\X}{\mathcal{X}}
\newcommand{\Y}{\mathcal{Y}}

\title{\quad 3D Hand Pose Estimation using \\\quad Simulation and Partial-Supervision \\\quad with a Shared Latent Space}

\addauthor{\quad Masoud Abdi}{mabdi@deakin.edu.au}{1}
\addauthor{\quad Ehsan Abbasnejad}{ehsan.abbasnejad@adelaide.edu.au}{2}
\addauthor{\quad Chee Peng Lim}{chee.lim@deakin.edu.au}{1}
\addauthor{\quad Saeid Nahavandi}{saeid.nahavandi@deakin.edu.au}{1}

\addinstitution{
 Institute for Intelligent System Research and Innovation (IISRI) \\
 Deakin University\\
 Waurn Ponds, VIC 3216, Australia
}
\addinstitution{
 Australian Centre for Visual Technologies (ACVT)\\
 University of Adelaide\\
 South Australia 5005, Australia
}

\runninghead{M. Abdi \etal}{Learning from Simulation and Partial-Supervision}

\def\etal{\emph{et al}\bmvaOneDot}

\begin{document}

\maketitle

\begin{abstract}
Tremendous amounts of expensive annotated data are a vital ingredient for state-of-the-art 3d hand pose estimation. Therefore, synthetic data has been popularized as annotations are automatically available. However, models trained only with synthetic samples do not generalize to real data, mainly due to the gap between the distribution of  synthetic and real data. In this paper, we propose a novel method that seeks to predict the 3d position of the hand using both synthetic and partially-labeled real data. Accordingly, we form a shared latent space between three modalities: synthetic depth image, real depth image, and pose. We demonstrate that by carefully learning  the shared latent space, we can find a regression model that is able to generalize to real data. As such, we show that our method produces accurate predictions in both semi-supervised and unsupervised settings. Additionally, the proposed model is capable of generating novel, meaningful, and consistent samples from all of the three domains. We evaluate our method qualitatively and quantitively on two highly competitive benchmarks (\textit{i.e.}, NYU and ICVL) and demonstrate its superiority over the state-of-the-art methods. The source code will be made available at \href{https://github.com/masabdi/LSPS}{https://github.com/masabdi/LSPS}.

\end{abstract}

%-------------------------------------------------------------------------
\section{Introduction}

Recovering the 3d configuration of the human hand has many applications, including augmented/virtual reality, human-computer interaction, sign language recognition, and robotics. Deep neural networks have obtained significant success in 3d hand pose estimation over the past few years~\cite{tompson2014real,oberweger2015hands}. These achievements, however, are highly dependent on the existence of massive amounts of training data, supervised by (usually) human-annotated targets, which are expensive and costly to acquire. As such, two extra sources of information have been extensively used to mitigate the need for the expensive annotations: synthetic data and unlabeled real data~\cite{wan2017crossing,spurr2018cross,wood2016learning,zimmermann2017learning,shrivastava2017learning,mueller2017ganerated}.

To this end, three main research directions have been proposed to tackle this shortcoming. The first line of work uses unlabeled real samples to learn representations that are useful for the semi-supervised learning task. As a result, they use the unlabeled samples along with a few labeled examples to obtain a more accurate regression model~\cite{wan2017crossing,spurr2018cross}. A second way to avoid the labeling burden is to use synthetic data, since annotations are automatically available~\cite{wood2016learning,zimmermann2017learning,abobakr2017rgb}. However, learning from synthetic data usually results in a sub-optimal solution. This is due to the \emph{domain-gap} between the distribution of synthetic and real data. Thus, the third direction of research incorporates both synthetic and unlabeled real data. It uses unlabeled real samples to map the synthetic data to a distribution close to that of the real data, and uses the annotations of the synthetic data to obtain a more accurate estimation of the hand pose~\cite{shrivastava2017learning,mueller2017ganerated}. % the distribution gapShrivastava~\etal~\cite{} tackles this problem by refining the synthetic data to match the distribution of the real data. Thus, the refiner network is trained to increases the realism of a given synthetic data, by competing against a discriminator whose task is to distinguish between the real and refined images.

In this paper, we introduce a novel 3d hand pose estimation method that uses both synthetic and partially-labeled real data. We dub it  Learning from Simulation and Partial-Supervision (LSPS). We formulate the problem as a generative modeling problem using a shared latent space. More specifically, we form a shared latent representation between: (1)~real depth image domain, (2)~synthetic depth image domain, and (3)~pose domain. Firstly, we show that by carefully learning the shared latent space using synthetic and unlabeled real samples, we can train a regressor that is able to generalize to real data. We then extend the proposed method to a semi-supervised setting, where we use real annotations to further enhance the performance of the model. Additionally, we show that samples from each of the three domains can be mapped to the latent space and back to the original form, and that they can be translated to other domains in a coherent and meaningful manner. The proposed model allows us to generate novel and consistent samples from all of the three domains. 

Our model based is on variational autoencoder (VAE)~\cite{vae} and generative adversarial network (GAN)~\cite{gan}. To obtain the shared latent representation, we use two VAE-GAN hybrids~\cite{aebp} to model the real and synthetic depth image domains, and a VAE for the pose domain. Since we do not generally have access to the corresponding real/synthetic image pairs, we exploit the weight-sharing and cycle-consistency constraints~\cite{liu2017unsupervised} to learn the mapping between the real and synthetic depth data. We also learn an auxiliary mapping function~$(M)$ and a posterior estimation function~$(P)$ that relate the latent space of the depth and that of the pose. This results in a single shared latent representation between the three domains.  Extensive experimental results on two real-world datasets, \textit{i.e.}, NYU~\cite{tompson2014real} and ICVL~\cite{tang2014latent}, demonstrate that the proposed method performs better than the existing models.

In summary, the \textbf{main contributions} of this work are as follows: \textbf{(1)} Presenting the first deep model (to the best of our knowledge) that learns from both synthetic and partially-labeled real data. \textbf{(2)} Proposing a novel shared latent space between three modalities, \textit{i.e.}, synthetic depth, real depth, and pose. \textbf{(3)} Performing extensive experiments on two highly competitive datasets and showing the superiority of our model over the state-of-the-art methods. 

\section{Related Work}\label{related}
 
3d hand pose estimation is a long-lasting problem in computer vision and related areas. It has received much attention recently due to its widespread application and affordable depth sensors~\cite{tang2014latent,sun2015cascaded,tompson2014real,zimmermann2017learning,oberweger2015training,oberweger2015hands,oberweger2017deepprior++,zhou2016model}. Approaches based on deep-learning, especially convolutional neural networks, are shown to be efficient and accurate in estimating the 3d position of the hand ~\cite{tompson2014real,oberweger2015hands,zimmermann2017learning,oberweger2015training,oberweger2017deepprior++,zhou2016model}. In this section, we outline some of the recent work surrounding our research.

\begin{figure}[t]
\begin{center}
\includegraphics[scale=.37]{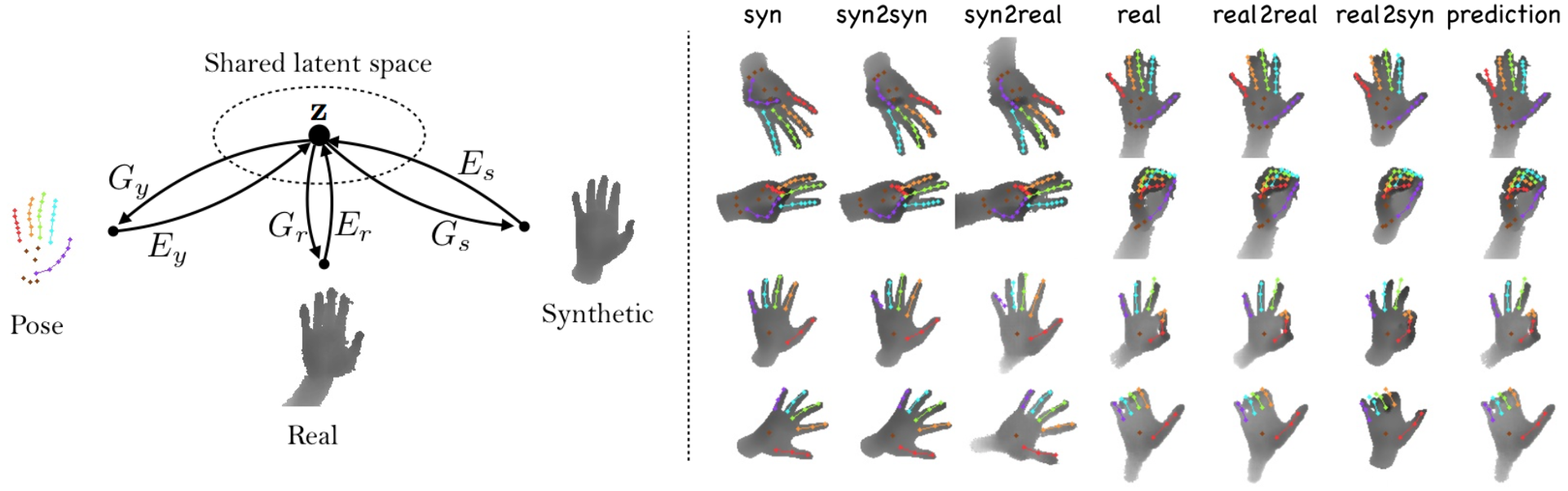}
\end{center}
\caption{\textbf{Left} depicts our shared latent variable assumption. The assumption implies the existence of a latent code $\bz$ for every given $(\bx_r,\bx_s,\by)$, such that samples from each domain can be encoded (decoded) to (from)  $\bz$ using proper encoding (decoding) functions $(E_r,E_s,E_y,G_r,G_s,G_y)$. \textbf{Right} shows transformations of the samples from one domain to other domains using the learned latent space.}
\label{fig:fig1}
\end{figure}

\textbf{Synthetic Data}. Synthetic data has been a popular choice for pose estimation~\cite{wood2016learning,zimmermann2017learning,tang2013real,shrivastava2017learning}. To reduce the domain-gap between synthetic and real data,  Tang~\etal~\cite{tang2013real} propose a transductive regression forest that uses unlabeled and synthetic data to estimate the 3d hand pose. Shrivastava~\etal~\cite{shrivastava2017learning} incorporate a refiner network that aims to improve the realism of the synthetic data through an adversarial training.  Recent work of Mueller~\etal~\cite{mueller2017ganerated} translates synthetic images to real data using a geometrically consistent image-to-image translation network, and predicts the position of the hand from RGB images. Rad~\etal~\cite{rad2017feature} use million-scale synthetic data, real data, and a feature matching strategy that is shown to produce high accuracy predictions.

\textbf{Generative Modeling}. Generative models, in particular VAEs~\cite{vae} and GANs~\cite{gan}, have recently shown to be very effective in many applications such as: image generation~\cite{vae,gan,improvedgan}, representation learning~\cite{infogan,afl}, and image-to-image translation~\cite{isola2017image,liu2017unsupervised}. Hybrid models combine multiple GANs and VAEs in order to take advantage of both frameworks~\cite{aebp,ali,liu2017unsupervised,afl}. Generative models have also proven to be efficient in 3d pose estimation from a single image~\cite{mueller2017ganerated,spurr2018cross}. 

\textbf{Shared Embedding}. The concept of shared embedding has been previously discussed in the literature, where it is assumed that samples from different modalities (\textit{i.e.} RGB image, depth image, pose, etc.) can be mapped to a shared embedding. Ngiam~\etal~\cite{ngiam2011multimodal} form a shared representation between audio and video. Ek~\etal~\cite{ek2007gaussian} and Navaratnam~\etal~\cite{navaratnam2007joint} use gaussian process latent variable models and form a shared latent representation between the image observations and human poses.  A cross-modal variational model derives a variational lower-bound based on VAE, that can be used to learn a shared latent space between different modalities~\cite{spurr2018cross}. Wan~\etal~\cite{wan2017crossing} model the shared latent variable using a combination of VAE and GAN and use a mapping function to relate the two latent spaces.

Our work is inspired by~\cite{wan2017crossing,liu2017unsupervised}. Liu~\etal~\cite{liu2017unsupervised} form a shared latent space between two image domains, and learn the mapping between the \emph{unpaired} samples using weight-sharing and cycle-consistency constraints. Wan~\etal~\cite{wan2017crossing} create a shared latent variable between depth image and pose domains using a combination of VAE and GAN. In this work, we use synthetic and partially-labeled real data to form a shared latent space between the two depth image domains (synthetic and real), and the pose domain. This allows us to find a regressor that performs efficiently on real data, even without any real annotations.

\begin{figure}[t]
\begin{center}
\includegraphics[scale=.4]{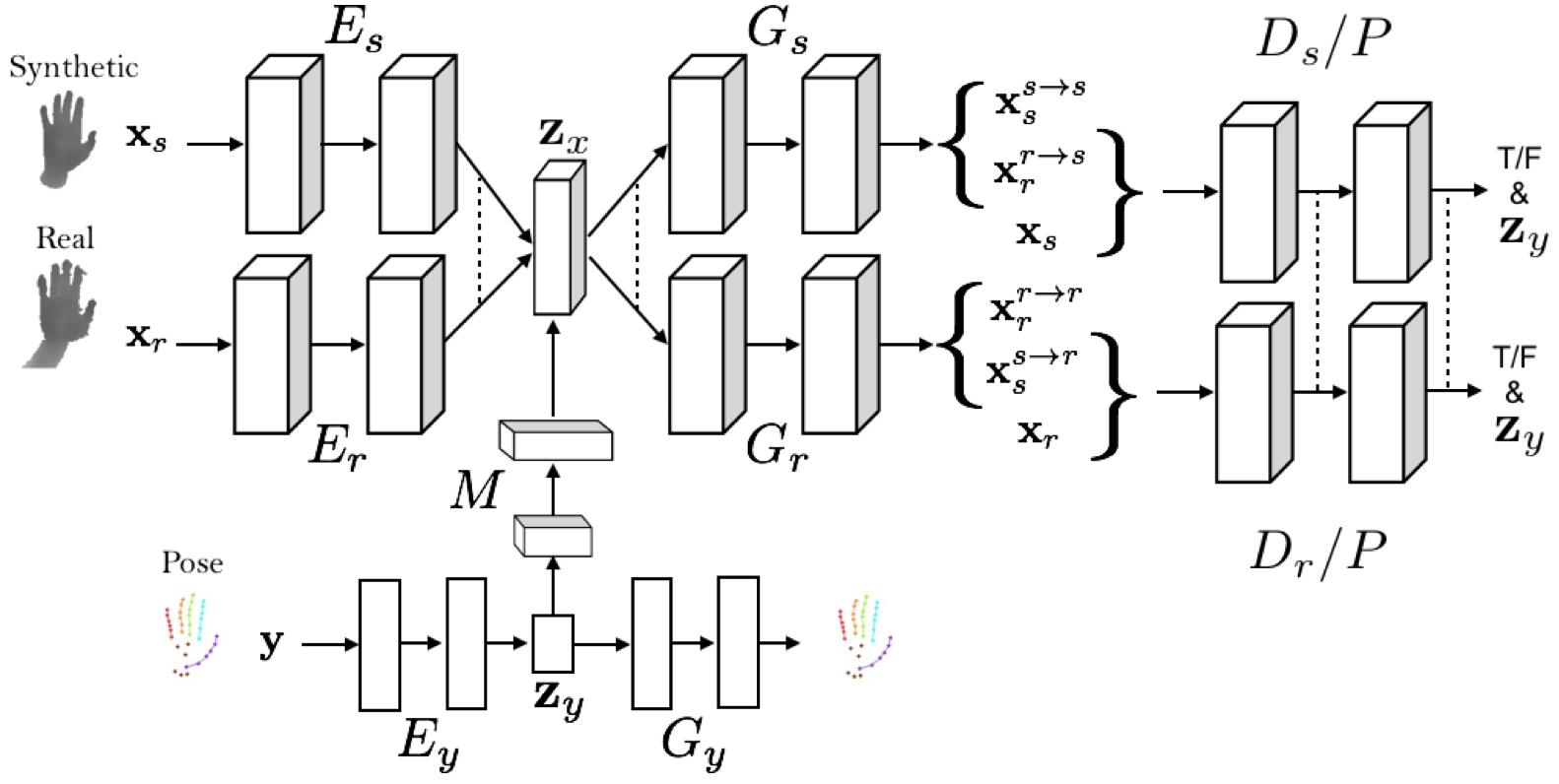}
\end{center}
\caption{\textbf{\small Visualization of the proposed model.} Cubes and rectangles represent convolutional and linear layers, respectively. Dotted lines denote weight-sharing. $\bz_y$ is the latent variable of the pose domain, while $\bz_x$ is the latent variable of the depth domain. }
\label{fig:vis}
\end{figure}

\section{Learning from Simulation and Partial-Supervision \small{(LSPS)}}

Let $\X_r, \X_s$, and  $\Y$ be the domain of real depth images, synthetic depth images, and poses, respectively. Using the shared latent space assumption, we assume that there exists a shared latent representation $\bz \in \mathbb{R}^d$ for every triplet $(\bx_r,\bx_s,\by)$ drawn from the joint distribution of the three domains $P_{\X_r, \X_s, \Y}(\bx_r,\bx_s,\by)$. As such, the latent code $\bz$ can be converted to (or recovered from) any of these three domains using proper encoding (or decoding) functions, see Figure~\ref{fig:fig1}. That is, we assume that we can find encoding ($E_r, E_s, E_y$) and decoding ($G_r, G_s, G_y$) functions such that all of these conditions hold: 
\begin{center}
$E_r(\bx_r)=E_s(\bx_s)=E_y(\by)= \bz, \quad G_r(\bz)=\bx_r$, $G_s(\bz)=\bx_s$, $G_y(\bz)=\by$.
\end{center}
Learning such latent representation is useful as it creates a shared understanding of the data in three domains, and enables us to transform samples from one domain to either of the other two domains. Furthermore, we can render the realistic depth map that corresponds to any arbitrarily hand pose $\by$ using the composition function $R_r = G_r(E_y(\by))$, or we can use the composite function of $P_r = G_y(E_r(\bx_r))$ to predict the 3d position of the hand from a real depth image.

In order to form this shared latent space, we use three components to model each of the three domains ($\X_r, \X_s, \Y$). The pose domain is modeled using a VAE as its low-dimensional embedding is shown to impose pose constraints that improve the reliability of the predictions~\cite{oberweger2015hands,wan2017crossing}. The depth image domains are modeled through VAE-GAN hybrids~\cite{aebp}, see Figure~\ref{fig:vis} for a visualization of the model.

\subsection{Pose Domain}

We first model the pose domain ($\text{VAE}_y$) using the variational upper bound as follows:
\begin{eqnarray}
\mathcal{L}_{\text{\tiny VAE}_y}(E_y,G_y)=&\lambda_0 \text{KL}( q_y(\bz_y|\by) || p(\bz_y) ) - \lambda_1 \mathbb{E}_{\bz_y \sim q_y (\bz_y|\by)}[\log p_{G_y}(\by|\bz_y)]
\label{eqn::vae_y}
\end{eqnarray}
where $q_y(\bz_y|\by)$ is our encoding distribution described by a multivariate Gaussian distribution $\mathcal{N}(E_{y;\mu}(\by), E_{y;\sigma^2}(\by))$ with mean $E_{y;\mu}(\by)$ and variance $E_{y;\sigma^2}(\by)$, and $p(\bz_y)$ is a Gaussian distribution with zero mean and unit variance $\mathcal{N}(0,I)$. We call this the \emph{pose latent space}. Throughout this paper, $\lambda_i$s are hyper-parameters that trade-off the relative importance of the terms. For example, here $\lambda_0$ and $\lambda_1$ control the relative importance of the KL and the negative log-likelihood term.

\subsection{The Shared Latent Space}

We then relate the two image domains by forming a shared latent space between them. Since we do not generally have access to the paired real/synthetic depth images, we use the weight-sharing and cycle-consistency constraints~\cite{liu2017unsupervised} to learn the shared latent space between the two image domains. We call this the \emph{depth latent space}. This is not sufficient for our formulation of the shared latent space, as we need to relate the pose latent space to the depth latent space. To this end, we define an auxiliary mapping function $M(\bz_y)$ that maps the samples of the pose latent space to those from the depth latent space~\cite{wan2017crossing}. 

To train the mapping function ($M$) we need the corresponding latent codes in the pose and the depth latent spaces. We obtain these codes using the synthetic pairs $(\bx_s, \by)$ and their respective encoding functions $(E_s, E_y)$. Therefore, we propose the following cost function to learn our shared latent space:
\begin{align}
\min_{M,P,E_s,E_r,G_s,G_r} \ \max_{D_s,D_r} \quad 
&\mathcal{L}_{\text{\tiny VAE}_s}(E_s,G_s) +\mathcal{L}_{\text{\tiny GAN}_s}(E_s,G_s,D_s) +\mathcal{L}_{\text{\tiny CC}_s}(E_s,G_s,E_r,G_r)\nonumber\\
+\ &\mathcal{L}_{\text{\tiny VAE}_r}(E_r,G_r) + \mathcal{L}_{\text{\tiny GAN}_r}(E_r,G_r,D_r)+\mathcal{L}_{\text{\tiny CC}_r}(E_r,G_r,E_s,G_s)\nonumber\\
+\ &\mathcal{L}_{\text{\tiny MAP}_s}(M,E_s,G_s,D_s)  + \mathcal{L}_{\text{\tiny POS}}(D_s,D_r,P) 
\label{eqn::UNIT_training},
\end{align}
where $\mathcal{L}_{\text{\tiny VAE}_s}, \mathcal{L}_{\text{\tiny VAE}_r}, \mathcal{L}_{\text{\tiny GAN}_s}, \mathcal{L}_{\text{\tiny GAN}_r}, \mathcal{L}_{\text{\tiny CC}_s}$ and $\mathcal{L}_{\text{\tiny CC}_r}$ are responsible for learning a shared latent space between the two depth domains, and $\mathcal{L}_{\text{\tiny MAP}_s}$ and $\mathcal{L}_{\text{\tiny POS}}$ align the latent space of the pose to that of the depth. We will discuss each of these terms in details, separately. 

\begin{figure}[t]
\begin{center}
\includegraphics[scale=.35]{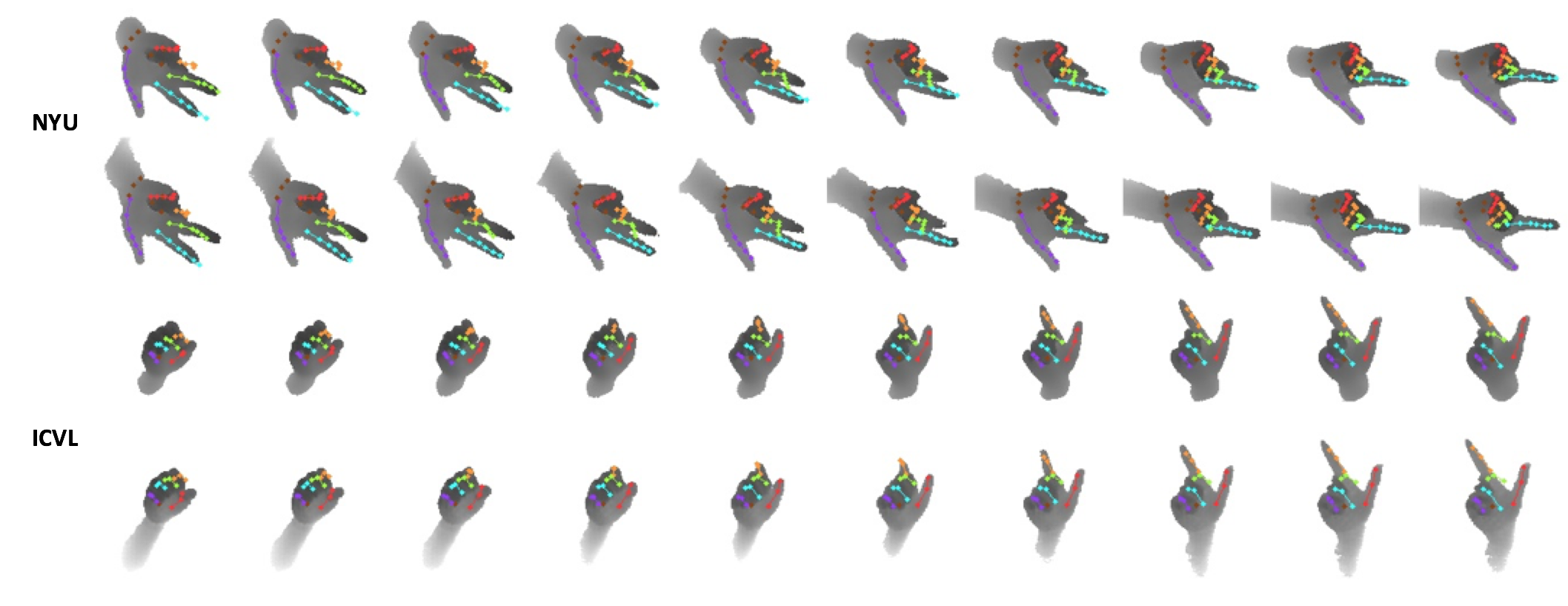}
\end{center}
\caption{\textbf{\small A random walk in the learned latent space.} First and second pairs of rows correspond to NYU and ICVL datasets, respectively. In each pair, the top row shows the generated synthetic samples and the bottom row shows the generated real samples. One can see that samples generated on the connecting line of two latent codes have a meaningful interpolation in all three (real, synthetic, and pose) domains.}
\label{fig:walk}
\end{figure}

The $\text{VAE}$ terms in Equation~\ref{eqn::UNIT_training} optimize a variational upper bound using the encoding functions, that output the mean $E_{s;\mu}(\bx_s)$ of a multivariate Gaussian distribution with unit variance $\big(q_s (\bz_s|\bx_s)\equiv\mathcal{N}(E_{s;\mu}(\bx_s), I)\big)$ as follows:
\begin{align}
\mathcal{L}_{\text{\tiny VAE}_s}(E_s,G_s)=&\lambda_2 \text{KL}( q_s(\bz_s|\bx_s) || p(\bz) ) - \lambda_3 \mathbb{E}_{\bz_s \sim q_s (\bz_s|\bx_s)}[\log p_{G_s}(\bx_s|\bz_s)],\label{eqn::vae_s}
\end{align}
where $p(\bz)\equiv\mathcal{N}(0,I)$, and $\mathcal{L}_{\text{\tiny VAE}_r}$ is defined in a similar way. The GAN terms in Equation~\ref{eqn::UNIT_training} make sure that the generated samples of $G_r$ and $G_s$ are realistic looking and indistinguishable from their domain, using the corresponding discriminators $D_r$ and $D_s$. Therefore, 
\begin{align}
\mathcal{L}_{\text{\tiny GAN}_s}&(E_s,G_s,D_s)=\lambda_4\mathbb{E}_{\bx_s \sim P_{\mathcal{X}_s}} [\log D_s (\bx_s) ]+\lambda_4\mathbb{E}_{\bz_r \sim q_r (\bz_r|\bx_r)}[\log(1-D_s(G_s(\bz_r)))].\label{eqn::gan_s}
\end{align}

Note that it suffices to apply the GAN cost function only to the translated images, as the reconstruction network is trained with the reconstruction term in Equation~\ref{eqn::vae_s}. The cycle-consistency term $\mathcal{L}_{\text{CC}_s}$ in Equation~\ref{eqn::UNIT_training} ensures that a twice-translated image resembles the initial image~\cite{liu2017unsupervised}. This is done using a VAE-like cost function as follows:
\begin{align}
\mathcal{L}_{\text{\tiny CC}_s}(E_s,G_s,E_r,G_r)=&\lambda_2\text{KL}( q_r(\bz_r|\bx_s^{s\rightarrow r})) || p(\bz) )-\lambda_3\mathbb{E}_{\bz_r \sim q_r (\bz_r|\bx_s^{s\rightarrow r})}[\log p_{G_s}(\bx_s|\bz_r)]\label{eqn::cc_s}.
\end{align} 
where $\bx_s^{s\rightarrow r}$ is the translated image from synthetic to real, which is obtained via $G_r(E_s(\bx_s))$.

We now relate the latent space of the pose and the latent space of the depth. We take the latent variable of the pose to be the reference variable, and learn a mapping function $M$ to the latent space of the depth as follows:
\begin{align}
\mathcal{L}_{\text{\tiny MAP}_s}(M,E_s,G_s,D_s)=&\ \lambda_5 \mathbb{E}_{\bz_y \sim q_y (\bz_y|\by), \bz_s \sim q_s (\bz_s|\bx_s)} || M(\bz_y) - \bz_s ||_2 \nonumber\\ 
+ & \ \lambda_6 \mathbb{E}_{\bz_y \sim q_y (\bz_y|\by)}[\log p_{G_s}(\bx_s|M(\bz_y))] \nonumber\\ 
+ & \ \lambda_4\mathbb{E}_{\bz_y \sim q_y (\bz_y|\by)}[\log(1-D_s(G_s(M(\bz_y))))]
\label{eqn::map}.
\end{align}
where $||.||_2$ is the 2-norm operator. The first term in Equation~\ref{eqn::map} ensures that samples from the latent variable of the pose are close to their corresponding latent codes when mapped through the mapping function $M$. The latent codes are obtained using synthetic data $(\bx_s, \by)$. The second and the third terms ensure that generation results of the mapped latent codes resemble their corresponding depth data.

Since the mapping function only operates in one direction (from the pose latent space to the depth latent space), we propose to estimate the posterior of the pose using a new function~$P$. The posterior estimation function $P: \mathcal{X}_s \rightarrow \bz_y$ aims to find the latent code for a given image in the pose latent space. Therefore,
\begin{align}
\mathcal{L}_{\text{\tiny POS}_s}(P)= &\ \lambda_{7} \mathbb{E}_{\bz_y \sim q_y (\bz_y|\by)}|| P(\bx_s) - \bz_y) ||_2
\label{eqn::pos}.
\end{align}
In practice, $P$ has the same architecture as the discriminators, thus it shares all of its layers with the discriminators, except for the last layer. Furthermore, the last term of our cost function is described as follows:
\begin{align}
\mathcal{L}_{\text{\tiny POS}}(D_s,D_r,P)=& \ \mathcal{L}_{\text{\tiny POS}_s}(P) +  \mathcal{L}_{\text{\tiny FM}}(D_s,D_r), \label{eqn::all}\\
\mathcal{L}_{\text{\tiny FM}}(D_s,D_r)= &\ \lambda_{8}||D_{s,\phi}(\bx_s^{s\rightarrow s}) - D_{r,\phi}(\bx_s^{s\rightarrow r})||_1 \nonumber\\ +  &\ \lambda_{8}||D_{s,\phi}(\bx_r^{r\rightarrow s}) - D_{r,\phi}(\bx_r^{r\rightarrow r})||_1 \label{eqn::fm},
\end{align}
where $D_{s,\phi}$ is the activations of the penultimate layer of the discriminator $(D_s)$, $\bx_s^{s\rightarrow s}$ and $\bx_r^{r\rightarrow r}$ are the reconstructed samples, while $\bx_s^{s\rightarrow r}$ and $ \bx_r^{r\rightarrow s}$ are the translated ones. The $\mathcal{L}_{\text{\tiny FM}}$ term is added to make sure that  discriminators, $D_s$ and $D_r$, similarly interpret the two corresponding samples generated using $G_s$ and $G_r$.

\subsection{Extension to Semi-Supervised Learning}\label{semi}

The flexibility of our model allows us to use any number of annotated real data to further enhance the overall performance of the model, when (partial-)supervision is available. This extra source of supervision is used to guide the mapping function $M$ to better align the two latent spaces of pose and depth. This is done by defining  $\mathcal{L}_{\text{\tiny MAP}_r}(M,E_r,G_r,D_r)$ term similar to  $\mathcal{L}_{\text{\tiny MAP}_s}$, and adding it to Equation~\ref{eqn::UNIT_training}. More importantly, we use the supervision of real data to approximate the posterior more accurately, by adding an extra term $\mathcal{L}_{\text{\tiny POS}_r}(P)$ to Equation~\ref{eqn::all}, where $\mathcal{L}_{\text{\tiny POS}_r}$ is also defined in a similar way to $\mathcal{L}_{\text{\tiny POS}_s}$.

\section{Implementation Details}

Our model consists of several sub-networks, some of which share a few layers (see Figure~\ref{fig:vis}). The image encoders and decoders $(E_s,E_r,G_s,G_r)$ have 3 convolutional layers and 4 residual blocks, where the first (last) residual block of the decoders (encoders) are shared. We use residual blocks with multiple residual connections \cite{xie2017aggregated,abdi2016multi}. The discriminators $(D_s,D_r)$ have 6 convolutional layers with 4 layers being shared between them. The pose model has one hidden layer with 30 units in both encoder and decoder $(E_y,G_y)$, and a 20 dimensional latent space. The mapping function ($M$) has four transposed convolutional layers. The posterior estimation function ($P$) is a network with 6 convolutional layers that shares all of its layers with the discriminators, except for the last layer. More details about the network architectures can be found in appendix.

We use Adam~\cite{adam} to train our model with a learning rate of $0.0001$. First, the pose model is trained for 200k iterations with a mini-batch size of 128 using Equation~\ref{eqn::vae_y}. The depth model and the mapping function are then trained (Equation~\ref{eqn::UNIT_training}) using the learned parameters of the pose model. Training continues for 500k iterations using one sample from each domain. Finally, the posterior estimation model is trained for 50k iterations with a mini-batch size of~32 (Equation~\ref{eqn::all}), resulting in our shared latent space. %Notice that we can use any fraction of real annotated data to train the model.
We note our design choice regarding the two latent spaces of depth and pose domains $(\bz_x, \bz_y)$, see Figure~\ref{fig:vis}. Although it is theoretically possible to have a single multi-modal latent space \cite{spurr2018cross}, we believe that having two connected latent spaces is of importance to the success of our method. This is because learning the mapping between synthetic and real data will be made easier and would require a less complicated network. As a result, generalization from synthetic to real data will be more effective by avoiding possible artifacts in down-sampling and/or up-sampling operations \cite{odena2016deconvolution}.

\section{Experiments}\label{exps}

\begin{figure}[t]
\begin{center}
\includegraphics[scale=.28]{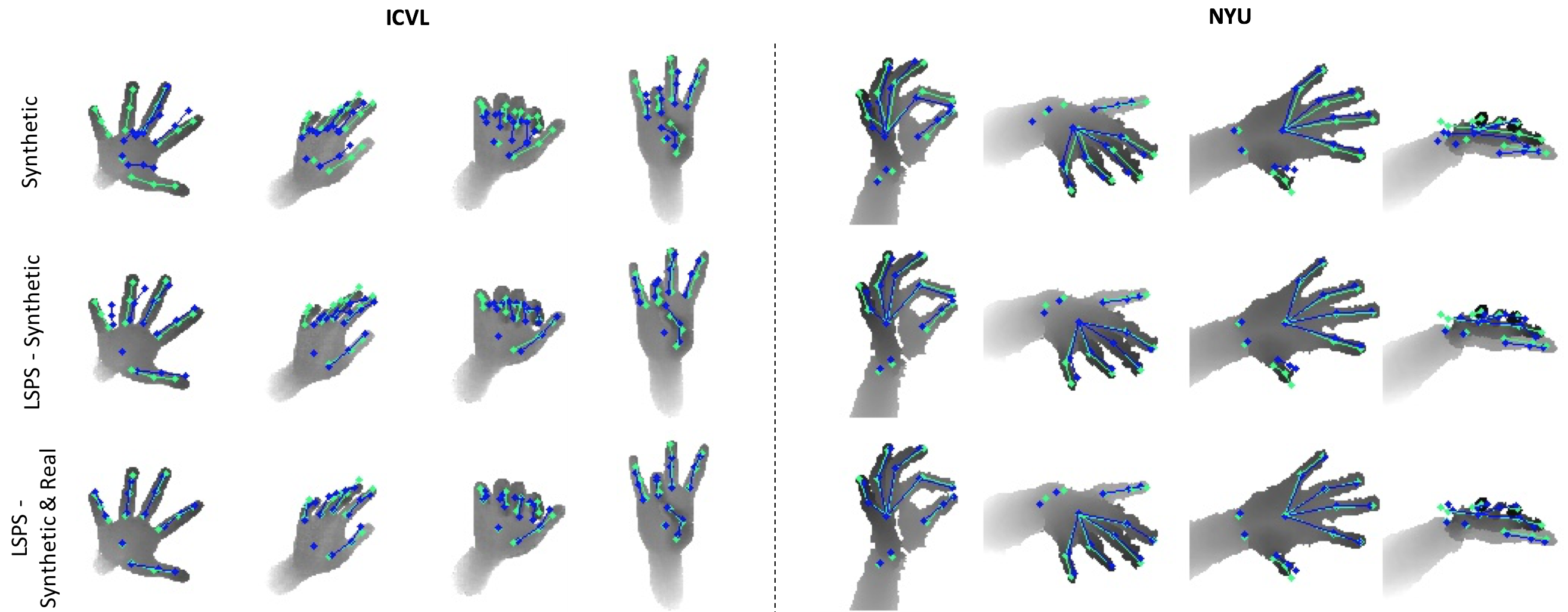}
\end{center}
\caption{\textbf{\small Qualitative evaluation of the proposed method.} Predictions and ground-truth annotations are depicted as blue and green, respectively. First row shows  the predictions of training only on the synthetic data. The second row depicts the predictions of LSPS, when only synthetic and unlabeled real data are used. The third row shows the predictions of LSPS using both synthetic and real data.}
\label{fig:prd}
\end{figure}

We evaluate our method qualitatively and quantitatively on two publicly-available datasets, \textit{i.e.}, NYU and ICVL. The NYU dataset~\cite{tompson2014real} contains very noisy depth images and has a wide range of poses, making it challenging for most pose estimation methods. It has around 72k samples for training and 8k for testing. The ICVL  dataset~\cite{tang2014latent} consists of 22k training and 1.6k testing samples with a large discrepancy between the training and testing sequences. We use the synthetic data of NYU with 72k depth images and their corresponding 3d poses~\cite{tompson2014real}. 

We follow the pre-processing pipeline of Oberweger~\etal~\cite{oberweger2017deepprior++}, where a fixed-size 3d cube centered at the center of the hand is cropped and resized to $128\times128$ depth images, normalized to $[-1,1]$. Since ICVL dataset contains left hand images, we flip the depth maps to resemble the right hand, as in the synthetic data. We augment the data with $180^\circ$ bidirectional rotation and $10mm$ translation. We observed that our model works with many choices of hyper-parameters. We set these hyper-parameters to be $\lambda_0=\lambda_2=0.1$, $\lambda_1=\lambda_3=\lambda_5=100$, $\lambda_4=\lambda_7=10$, $\lambda_6=10000$, and $\lambda_{8}=1$. We also consider  Equation~\ref{eqn::fm} with $\lambda_{8}=0.0001$ when training the depth model using Equation~\ref{eqn::UNIT_training}.

\begin{table}[ht]
\begin{center}
\begin{tabular}{l|cc||cc}
  & \multicolumn{2}{c}{\bf Joint mean error (mm)} & \multicolumn{2}{c}{\bf \% of frames within 40mm} \\
  & \bf{NYU} & \bf{ICVL} & \bf{NYU} & \bf{ICVL} \\
\hline 
Synthetic-Only &   25.24   &  23.96  & 36.57 & 51.07   \\ 
Synthetic-Only* &   23.71  &   23.91  & 43.60   & 48.37  \\ 
LSPS-Synthetic [Ours] &   \textbf{17.84}  &   \textbf{14.09}  &   \textbf{62.57}  & \textbf{86.78}\\ 
\hline 
Real data* &   15.83  &   7.09  &  73.62 & 96.18   \\ 
\end{tabular}
\caption{\textbf{\small Comparison of our method with two baselines on NYU and ICVL}. One can see that our model performs significantly better than the models trained only on the synthetic data, by incorporating unlabeled real samples. The rows with * use our pre-trained model.}
\label{table:sim}
\end{center}
\end{table}

\textbf{Learning from simulation and unlabeled real data}. We demonstrate that effective learning from unlabeled real data and synthetic data can lead to enhanced generalization. We compare our method with a model trained only on the synthetic data. We also use  our pre-trained model and train a second baseline on the synthetic data to have a fair comparison. We use two metrics in our evaluation: joint mean error (in mm) averaged over all joints and all frames, and percentage of  frames in which all joints are below a certain threshold~$(d)$ \cite{taylor2012vitruvian}. Following the previous work~\cite{oberweger2015hands,oberweger2015training}, we only evaluate 14 joints from the NYU dataset.

\begin{figure}[t]
\begin{center}
\includegraphics[scale=.52]{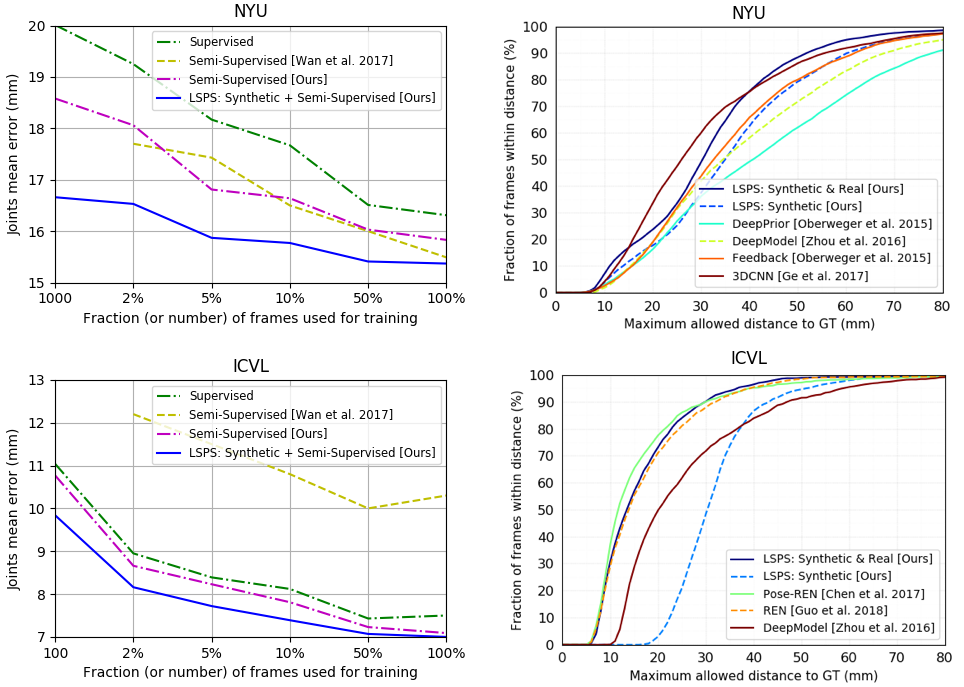}
\end{center}
\caption{\textbf{\small Comparison to the state-of-the-art hand pose estimation models.} \textbf{Left} compares against the existing semi-supervised method. Our model uses synthetic and partially-labeled real data to predict the hand pose more accurately. This is particularly more apparent when fewer real annotations are available. Note that \textit{x-axis} is not linear in scale. \textbf{Right} compares our model against existing fully-supervised methods. The proposed model performs comparable or better than the existing methods. It can be seen that our method outperforms some of the most recent state-of-the-art methods in an unsupervised setting.}
\label{fig:curves}
\end{figure}

Table \ref{table:sim} shows the results of the experiments against the two baselines.  The proposed method achieves a mean error of \textbf{\small17.84mm} on NYU dataset, whereas training only on the synthetic data obtains an error rate of \textbf{\small23.71mm}. It can be seen that our result is quite close to training on the real data with \textbf{\small15.83mm} error. This demonstrates the benefits of learning from synthetic and unlabeled real data. On ICVL dataset, our method obtains \textbf{\small14.09mm} mean error, while training on the synthetic data obtains a poor generalization performance of \textbf{\small23.91mm}. Nevertheless, our method outperforms the baselines with about \textbf{19\%} and \textbf{35\%} absolute improvement in terms of the number of frames within 40mm on NYU and ICVL, respectively. Additionally, experiments on ICVL dataset demonstrate that our method is capable of learning from very different domains, where there exists a large discrepancy between the synthetic and real data.

\textbf{Using partial supervision}. We show that using a few labeled examples can significantly improve the performance of our proposed model. We uniformly sample $m\%$ of the real data and use their corresponding annotations during the training. As such, we also consider the cost functions described in Section \ref{semi} when training the models. We compare our results with the state-of-the-art hand pose estimation methods (see Figure~\ref{fig:curves}). On the left we compare  against crossingNet~\cite{wan2017crossing}. We also train our version of \cite{wan2017crossing} using the same network to have a fair comparison. Our method shows up to $17\%$ and  $10\%$ relative improvement on NYU as compared to the supervised baseline and the semi-supervised method of crossingNet~\cite{wan2017crossing}, respectively. On ICVL dataset $11\%$ and  $9\%$ improvement was obtained relative to the two other methods. The proposed method consistently outperforms the other two methods, especially when fewer labels are available. On the right (Figure~\ref{fig:curves}), we compare our method with the existing fully-supervised methods \textit{i.e.}, DeepPrior\cite{oberweger2015hands}, DeepModel\cite{zhou2016model}, Feedback\cite{oberweger2015training}, 3DCNN\cite{ge20173d}, REN\cite{guo2017towards}, and Pose-REN\cite{chen2017pose}.  Our method produces accuracies comparable or better than the existing methods in the fully-supervised experiments.

\textbf{Generative capabilities}. As outlined earlier, our method is capable of generating meaningful samples from all of the three domains. Figure~\ref{fig:fig1} (right) shows that samples can be encoded to the latent space and decoded to any of the three domains, including their original domain. We also show the smoothness of the shared latent space by synthesizing samples along the connecting line of two latent codes (see Figure~\ref{fig:walk}). This demonstrates that the learned latent space represents a valid statistical \emph{multi-modal} model of the human hand. Interestingly, samples from the prior distribution $\mathcal{N}(0,I)$ can be used to generate valid data in all three domains. The generative capabilities of the proposed model can potentially be used to produce labelled data to further enhance the performance of existing models.

\textbf{Qualitative results}. We show the discriminative results of our proposed model in Figure~\ref{fig:prd}. The first row shows the predictions of a model trained only on  the synthetic data. The second row shows our predictions when only unlabeled real data and synthetic data are used, and the last row uses both synthetic and real data. One can see that our method produces more accurate predictions of the hand pose as compared to the model trained only on the synthetic data. See appendix for more qualitative results and some failure cases. We noticed that in some parts of the real distribution, LSPS-synthetic does not significantly improve the predictions compared to the synthetic-only case. We believe that this is due to the sparsity of the real and/or synthetic data in those parts of the distribution. Nevertheless, using real annotations along with synthetic samples consistently improves the quality of the predictions.

\section{Conclusion}
In this paper, we tackle the problem of 3d hand pose estimation using synthetic and partially-labeled real data. We form a shared latent space between real depth, synthetic depth, and pose domains. We use several techniques including the cycle-consistency and weight-sharing constraints to learn the shared latent space using  unpaired real/synthetic samples. We show that the shared latent space facilitates a very accurate model that is able to generalize from synthetic data to real data. We also demonstrate that synthetic data can enhance the robustness of the model when partial supervision is available. The shared latent space allows us to generate samples consistently from all of the three domains. %Empirical evaluations on two challenging datasets indicate the superiority of our method compared to the state-of-the-art models.

\bibliography{egbib}

\section{APPENDIX}
\subsection{Network Architectures}

The architecture of the pose model is represented in Table~\ref{tbl::arch_vae}. The input to the pose encoder is a vector of $3J$ real numbers corresponding to $J$ coordinates ($J=36$ for NYU, and $J=16$ for ICVL).  The depth model uses convolutional layers and residual blocks \cite{abdi2016multi,xie2017aggregated} with instance normalization. LeakyReLU activation function is used in almost of the layers of the model (see Table~\ref{tbl::arch_depth}). The mapping function $M$ consists of four transposed-convolutional layers that converts a $20$ dimensional vector to the latent space of the depth with $32\times32$ spatial dimensions, see Table \ref{tbl::arch_map}.

\begin{table}[tbh!]
\small
\centering
\begin{tabular}{cl|cl}
\hline
Layer & Encoder: $E_y$ & Layer & Decoder: $G_y$\\\hline
0 & INPUT-(3J) & 0 & INPUT-(20)\\
1 & FC-(30N), LeakyReLU & 1 & FC-(30N), LeakyReLU\\
$\mu$ & FC-(20N) & 2 & FC-(3J)\\
$\sigma^2$ & FC-(20N)\\\hline
\end{tabular}
\caption{\small Network architecture of the pose model.}\label{tbl::arch_vae}
\end{table}

\begin{table}[tbh!]
\small
\centering
\begin{tabular}{clc}
\hline
Layer &  Mapping: $M$  \\\hline
0 & INPUT-(3J)\\
1 & TCONV-(N1024,K4,S1), LeakyReLU\\
2 & TCONV-(N1024,K4,S2), LeakyReLU\\
3 & TCONV-(N512,K4,S2), LeakyReLU\\
4 & TCONV-(N256,K4,S2)\\
\end{tabular}
\caption{\small Network architecture of the mapping function.}\label{tbl::arch_map}
\end{table}

\begin{table}[tbh!]
\small
\centering
\begin{tabular}{clcc}
\hline
Layer &  Encoders: $E_s,E_r$ & Shared? & \\\hline
1 & CONV-(N64,K7,S1), LeakyReLU &\xmark\\
2 & CONV-(N128,K3,S2), LeakyReLU  &\xmark\\
3 & CONV-(N256,K3,S2), LeakyReLU  &\xmark\\
4 & RESBLK-(N512,C4,K3,S1) &\xmark\\
5 & RESBLK-(N512,C4,K3,S1) &\xmark\\
6 & RESBLK-(N512,C4,K3,S1) &\xmark\\
$\mu$ & RESBLK-(N512,C4,K3,S1) &\cmark\\\hline
Layer &  Generators: $G_s,G_r$ & Shared?\\\hline
1 & RESBLK-(N512,C4,K3,S1) &\cmark\\
2 & RESBLK-(N512,C4,K3,S1) &\xmark\\
3 & RESBLK-(N512,C4,K3,S1) &\xmark\\
4 & RESBLK-(N512,C4,K3,S1) &\xmark\\
5 &  TCONV-(N256,K3,S2), LeakyReLU &\xmark\\
6 &  TCONV-(N128,K3,S2), LeakyReLU &\xmark\\
7 &  TCONV-(N3,K1,S1), TanH & \xmark\\\hline
Layer &  Discriminators: $D_s,D_r$ & Shared? \\\hline
1 & CONV-(N64,K3,S2), LeakyReLU &\xmark\\
2 & CONV-(N128,K3,S2), LeakyReLU  &\xmark\\
3 & CONV-(N256,K3,S2), LeakyReLU  &\cmark\\
4 & CONV-(N512,K3,S2), LeakyReLU  &\cmark\\
5 & CONV-(N1024,K3,S2), LeakyReLU  &\cmark\\
6 & CONV-(N1,K2,S1), Sigmoid &\cmark\\\hline
Layer &  Posterior: $P$ & Shared with $D_s$, $D_r$? \\\hline
1 & CONV-(N64,K3,S2), LeakyReLU &\cmark\\
2 & CONV-(N128,K3,S2), LeakyReLU  &\cmark\\
3 & CONV-(N256,K3,S2), LeakyReLU  &\cmark\\
4 & CONV-(N512,K3,S2), LeakyReLU  &\cmark\\
5 & CONV-(N1024,K3,S2), LeakyReLU  &\cmark\\
6 & CONV-(N20,K2,S1) &\xmark\\\hline
\end{tabular}
\caption{\small Network architecture of the depth model.}\label{tbl::arch_depth}
\end{table}
\newpage

\subsection{More Qualitative Results and Failure Cases}

\begin{figure}[h]
\begin{center}
\includegraphics[scale=.4]{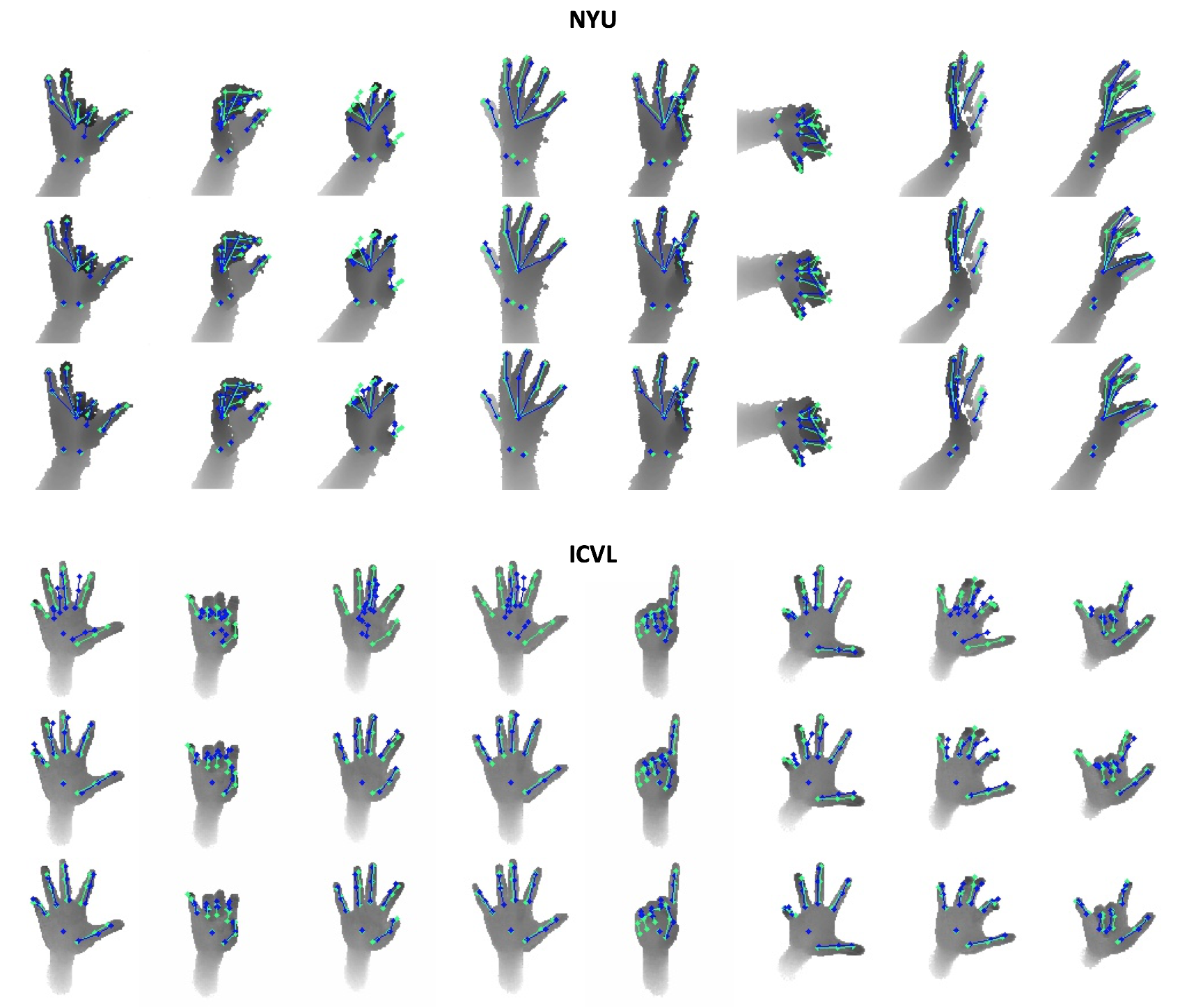}
\end{center}
\caption{More discriminative results on NYU and ICVL.}
\label{fig:app1}
\end{figure}

\begin{figure}[h]
\begin{center}
\includegraphics[scale=.4]{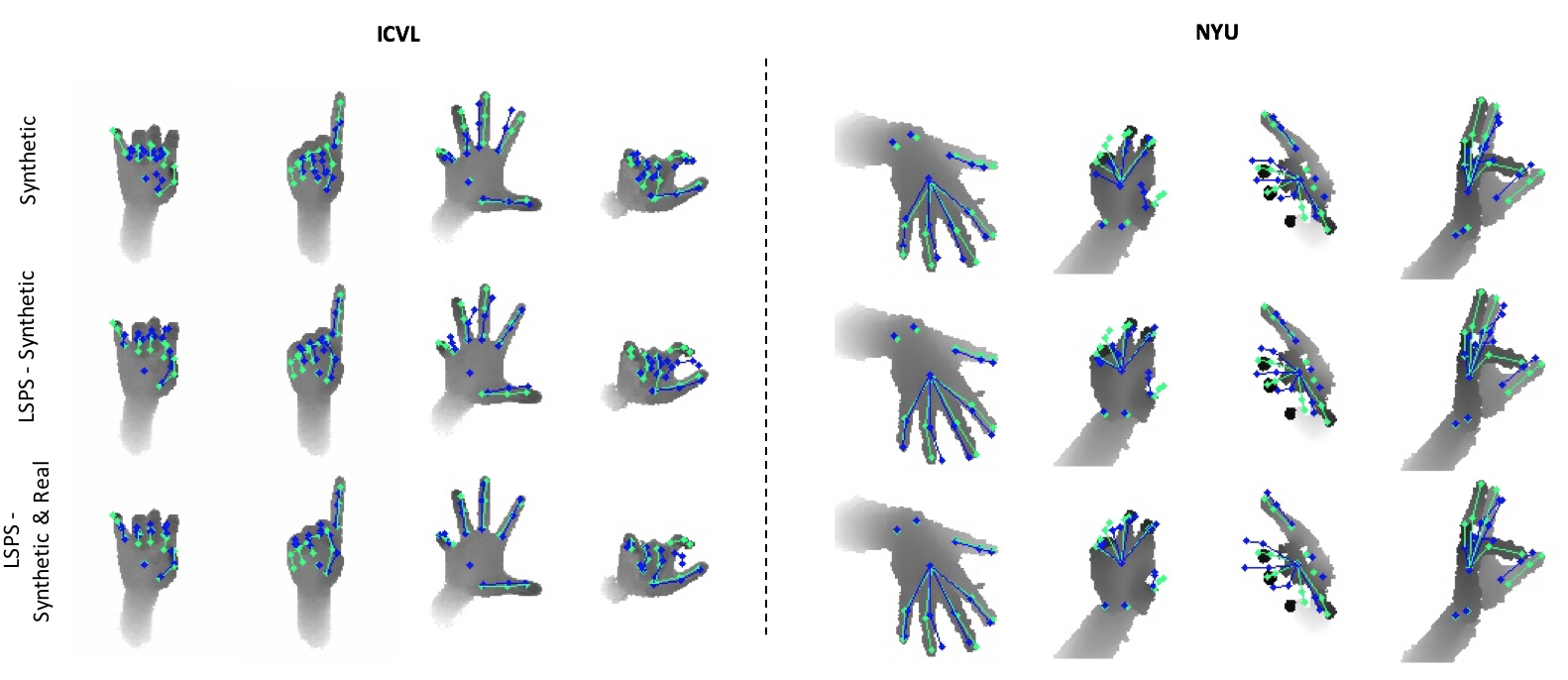}
\end{center}
\caption{Some failure cases. In some parts of the real distribution, LSPS-synthetic does not necessarily improve the predictions as compared to the synthetic-only case. However, using real annotations along with the synthetic samples consistently improve the quality of the predictions.}
\label{fig:app3}
\end{figure}

\begin{figure}[h]
\begin{center}
\includegraphics[scale=.4]{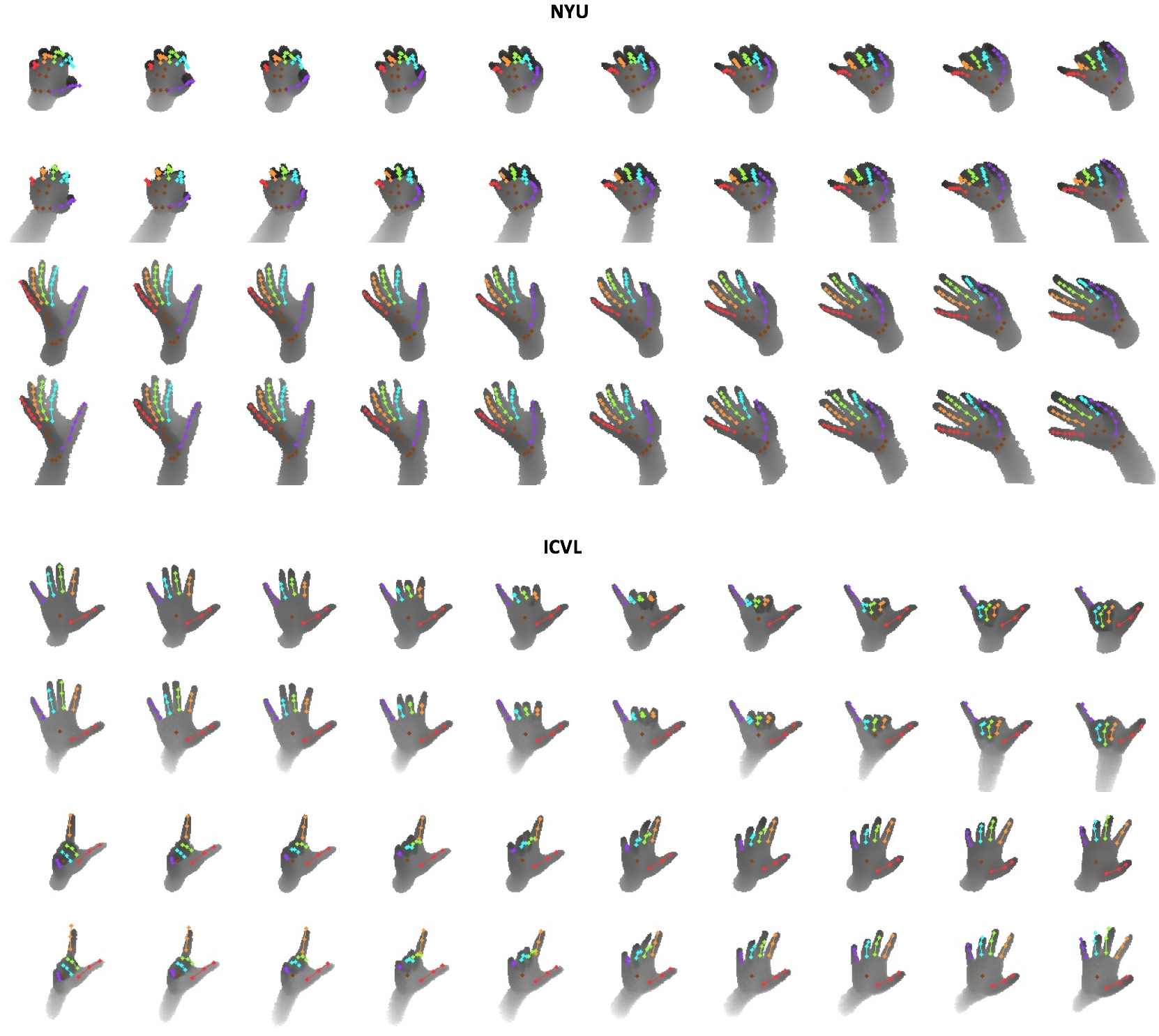}
\end{center}
\caption{More generative results on NYU and ICVL.}
\label{fig:app2}
\end{figure}

Figure~\ref{fig:app1} depicts the discriminative results and Figure~\ref{fig:app2} show the generative results. Also, some failure cases are shown in Figure~\ref{fig:app3}. 

\end{document}